\documentclass{article}
\usepackage{spconf,amsmath,graphicx}
\usepackage{graphicx}
\usepackage{amsmath}
\usepackage{multirow}
\usepackage{amssymb}
\usepackage{xcolor}
\usepackage{dsfont}
\usepackage{booktabs}

\usepackage{algorithm}
\usepackage{algpseudocode}

\usepackage{hyperref}
\hypersetup{
	colorlinks=true,
	linkcolor=blue,
	filecolor=magenta,      
	urlcolor=blue,
	pdfpagemode=FullScreen,
}

\usepackage{cleveref}

\usepackage{fancyhdr}
\fancypagestyle{firstpage}{
	\fancyhf{}
	\fancyfoot[L]{\footnotesize \copyright 2025 IEEE. Personal use of this material is permitted. Permission from IEEE must be obtained for all other uses, in any current or future media, including reprinting/republishing this material for advertising or promotional purposes, creating new collective works, for resale or redistribution to servers or lists, or reuse of any copyrighted component of this work in other works.}

}

\title{{LINEA}: Fast and Accurate Line Detection Using Scalable Transformers}
\name{Sebastian Janampa and Marios Pattichis}
\address{The University of New Mexico\\Dept. of Electrical \& Computer Engineering}

\begin{document}
	%
	\maketitle
	\thispagestyle{firstpage}
	
	\begin{abstract}
Line detection is a basic digital image processing operation used by higher-level processing methods. Recently, transformer-based methods for line detection have proven to be more accurate than methods based on CNNs, at the expense of significantly lower inference speeds. As a result, video analysis methods that require low latencies cannot benefit from current transformer-based methods for line detection. In addition, current transformer-based models require pretraining attention mechanisms on large datasets (e.g., COCO or Object360). This paper develops a new transformer-based method that is significantly faster without requiring pretraining the attention mechanism on large datasets. We eliminate the need to pre-train the attention mechanism using a new mechanism, Deformable Line Attention (DLA). We use the term LINEA to refer to our new transformer-based method based on DLA. Extensive experiments show that LINEA is significantly faster and outperforms previous models on sAP in out-of-distribution dataset testing. Code available at \url{https://github.com/SebastianJanampa/LINEA}.
	\end{abstract}
	
	
	\section{Introduction}
	Line segment detection is a low-level operation in digital image processing.
	Successful line segment detection is essential for higher-order operations such as camera calibration,
	scene understanding, SLAM, and as a low-level feature for self-driving vehicles.
	Thus, it is clear that the development of fast and accurate line (or line segment) detection can impact several applications.
	
	Most modern methods use convolutional neural networks (CNN) to detect lines.
	We summarize the CNN architectures for line detection as a sequence of basic steps.
	First, they use a light encoder-decoder backbone to generate a single feature map, 
	followed by a module to produce a non-fixed number of line candidates \cite{hawp}. 
	Next, a refinement module receives the candidates to produce a more accurate estimation of the lines. 
	Then, a Line of Interest (LoI) \cite{lcnn} module predicts whether the refined candidate is a line.

	
	End-to-end transformer models \cite{detr} consist of a backbone that produces a set of feature maps and 
	an encoder to enhance them. Then, a decoder uses a fixed number of queries to estimate the line endpoints and 
	the probability that a query is associated with an actual line segment. 
	LinE segment TRansform (LETR) \cite{LETR} is a transformer model that uses a coarse-to-fine approach to detect lines.
	More recently, the Deformable Transformer-based Line Segment Detector (DT-LSD)  \cite{dtlsd} uses a deformable 
	attention mechanism and the Line Contrastive DeNoising (LCDN) technique to accelerate convergence during training.
	
	
	Ultimately, we are interested in attention mechanisms that do not require expensive pretraining on large datasets. We introduce
	a new deformable line attention mechanism (DLA) to eliminate the need for pre-training.
	
	We introduce a family of transformer-based models termed LINEA  (Spanish word for line) that can detect line segments
	with constant inference latency. LINEA models use the new deformable line attention (DLA) to achieve competitive results. 
	Our approach uses the minimum number of trainable parameters compared to other transformer-based methods. 
	In out-of-distribution testing, we show that LINEA models give competitive or better results on the YorkUrban dataset.
	Furthermore, our three smaller LINEA models require fewer FLOPs than all models.
	
		\begin{figure*}[t]
		\centering
		\includegraphics[width=0.95\linewidth]{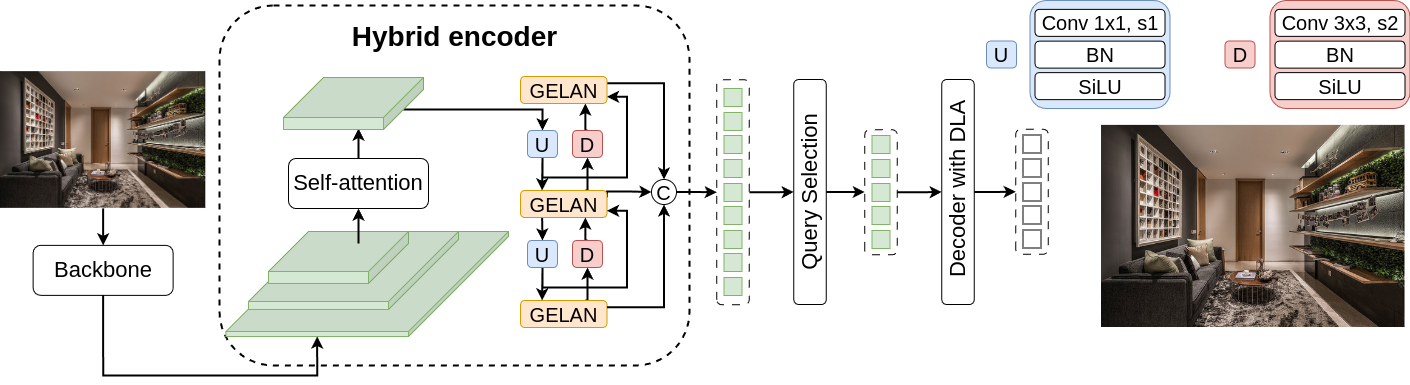}
		\caption{Architecture of LINEA. The model has a hierarchical backbone that extracts a set of feature maps from the input imagee. A modified GELAN fuses two feature maps to give them global and precise information. The following module selects the queries with the highest probabilities of storing a line candidate. Then, the decoder uses the deformable line attention (DLA) mechanism to make the queries and the feature maps interact to predict the line's endpoints.}
		\label{fig:architecture}
	\end{figure*}
	
	\section{Background}
	
	\subsection{CNNs for line detection}	The majority of deep learning methods for line detection are based on convolutional neural networks (CNNs).
	L-CNN \cite{lcnn} uses a light encoder-decoder encoder that processes RGB images to produce a feature map 
	whose dimensions are a fourth of the original image size. 
	A proposal initialization module uses a linear layer to produce junction candidates from the feature map, 
	followed by non-maximum suppression to remove duplicates around correct predictions and generate a non-fixed number 
	of line candidates from the junctions. Then, a Line of Interest (LoI) module determines whether a line candidate is an actual line.
	HAWP \cite{hawp} and HAWPv2 \cite{HAWPv2} reduce the number of line candidates the LoI module receives by using a 
	4D attraction field and holistic attraction field, respectively. SACWP \cite{sacwp} improves the backbone's decoder from 
	HAWP by using asymmetric convolutions. MLNET \cite{mlnet} adds enhancing blocks 
	based on the self-attention \cite{attention} mechanism to the backbone's encoder from HAWP. ULSD \cite{ulsd} uses 
	Bezier curve representations to handle different types of images. F-CLIP \cite{fclip} is a single-stage network 
	with a light encoder-decoder backbone followed by specialized heads that estimate the line center, length, and angle.

	\subsection{Transformer-based methods for line detection}         
	LinE segment TRansformers (LETR) \cite{LETR} is the first end-to-end line detector that uses transformers. 
LETR uses the attention mechanism to detect line endpoints. 
However, LETR has high latency and is slow to train (requiring $>800$ epochs for training).
LETR also requires significant memory resources.
More recently, Deformable Transformer-based Line Segment Detection (DT-LSD) \cite{dtlsd} uses 
a deformable attention mechanism and a contrastive denoising technique for line detection.
Compared to the methods proposed in the current paper, we note that both methods require training
on large datasets (e.g., Object365 or COCO).
	
	\section{Methodology}
	
	\Cref{fig:architecture} illustrates the general architecture of our proposed model. It has a hierarchical backbone that generates a set of feature maps. Then, the hybrid encoder enhances the feature maps. Next, the query selection module selects the most important pixels of the feature maps to initialize the queries. Finally, the decoder and the DLA mechanism use the queries and enhanced feature maps to estimate the lines' endpoints.

	
	\subsection{Hybrid encoder}
	A set of $1\times1$ convolutions ensures that all feature maps share the number of channels. Then, self-attention enhances the smallest spatial feature map. Next, a modified GELAN module fuses two adjacent feature maps for cross-scale interaction. The modified GELAN uses two parallel $1\times3$ and $3\times1$  convolutional layers to extract better line features and the  $3\times3$ and $1\times1$ convolutions. We fuse all layers into one $3\times3$ kernel for deployment to avoid increasing the latency.
	
	\subsection{Query selection}
	A linear layer processes the encoder's output to get the top-$k$ pixels with the highest probability of containing an instance. 
	In what follows, similar to transformer frameworks for object detection, we refer to the line's endpoints as an anchor box.
	The initial anchor boxes are the normalized coordinates of the selected pixel on its feature map plus the offsets 
	generated by another linear layer that processes the top-$k$ pixels. We refer to the anchor boxes as static positional
	queries to highlight the fact that they are not associated with learnable parameters from the model. 
	\Cref{alg:anchor_generation} provides $\mathrm{pytorch}$-styled pseudocode for the entire process.
	In contrast, we note that the content queries are learnable because they are embeddings that are updated after each iteration during training.
	
	\begin{algorithm}[!t]
		\caption{Anchor generation}\label{alg:anchor_generation}
		\begin{algorithmic}
			\Require $\mathbf{F}$. Concatenation of multiple flattened feature maps.
			\Require $k$. Number of queries
			\Require lvl\_shape. A list with the dimensions of each feature map in $\mathbf{F}$
			
			\State
			\State $\text{prob embeddings} \gets \mathrm{LinearLayer}(\mathbf{F})$
			
			\State $\text{proposals}, \text{indices} \gets \mathrm{torch.topk}(\text{prob embeddings, }k)$ 
			
			\State $\text{coords} \gets \mathrm{get\_norm\_coords}(\text{indices, lvl\_shape})$ 	
			
			\State $\text{coords} \gets \mathrm{inverse\_sigmoid}(\text{coords})$ \Comment{coords $\in \mathbb{R}^{k\times2}$}
			
			\State $\text{offsets} \gets \mathrm{LinearLayer}(\text{proposals})$ \Comment{offset $\in \mathbb{R}^{k\times4}$}
			
			\State $\text{anchors} \gets \text{coords}.\mathrm{repeat}(1, 2) + \text{offsets}$
			
			\Return anchors
		\end{algorithmic}
	\end{algorithm}
	
	\subsection{Decoder}
	
	The decoder uses D-FINE heads and its fine-grained distribution refinement \cite{dfine} to estimate the line's endpoints from the enhanced feature maps and the top-$k$ queries.
	
	%

	\subsubsection{Deformable line attention}
	
\begin{figure}[!t]
		\centering
		\includegraphics[width=\columnwidth]{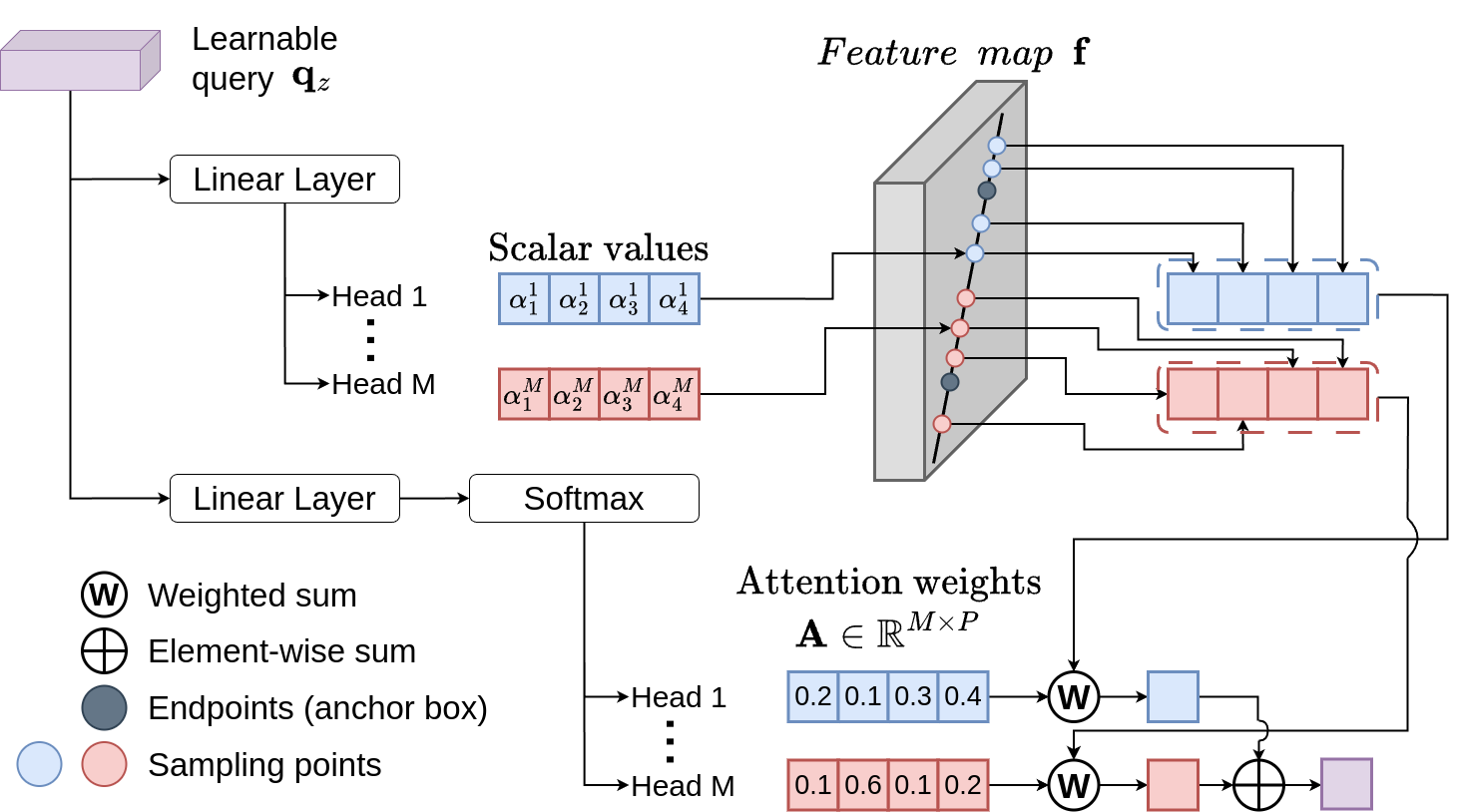}
		\caption{Deformable line attention. Two linear transformations generate scalar values $\alpha$ and the attention matrix $\mathbf{A}$  from the learnable query $\mathbf{q}_z$. DLA has a single ($L=1$) feature map and samples $P=4$ points in the example. A property of DLA is that all sampling points align with the endpoints.}
		\label{fig:deformable_line_attn}
	\end{figure}

	\textit{Deformable line attention} (DLA) mechanism
	is a new method designed for lines which, unlike regular objects, do not have area. Our approach takes motivation from the deformable attention \cite{deformable-DETR} used for object detection. We provide a top-level diagram of our approach on \cref{fig:deformable_line_attn}
	

We begin with the candidate endpoints $\mathbf{ep}_1$ and $\mathbf{ep}_2$.
We represent the vector query $\mathbf{q}_z$ using $\mathbf{q}_z=(\mathbf{ep}_1, \mathbf{ep}_2)$.
Here, we note that the midpoint between the two endpoints is given by $(\mathbf{ep}_1 + \mathbf{ep}_2)/2$
while the vector difference between them is given by $\Delta \mathbf{ep}= \mathbf{ep}_1 - \mathbf{ep}_2$.
For each feature map, denoted as $\mathbf{f}$ in \cref{fig:deformable_line_attn}, 
DLA computes sampling points $\mathbf{p}_i$ given by:
\begin{equation}
	\mathbf{p}_i = \alpha_i\Delta \mathbf{ep} + \frac{\mathbf{ep}_1 + \mathbf{ep}_2}{2}, \text{\phantom{aa}for $i=1, 2,\ldots,  LPM$}
	\label{eq:sampling}
\end{equation}
where the $\alpha_i$ represent learnable steplengths associated with each attention head,
$P$ represents the number of sampling points per feature map, 
$L$ represents the number of feature maps,
$M$ represents the number of attention maps,
and hence $L P M$ represents the total number of learnable steplengths
(see \cref{fig:deformable_line_attn}).
To facilitate fast memory access, the sampling points are stored in a single matrix     
$\mathbf{S} \in \mathbb{R}^{M\times L\times P \times 2}$. 

We let $\mathbf{A} \in \mathbb{R}^{M\times L \times P}$ denote the attention matrix.
We use the softmax function to normalize each attention map as given by:
$\sum_{l=1}^L\sum_{p}^{P}\mathbf{A}_{mlp} = 1$ for $m=1,2\ldots, M$. 
We thus have that the DLA mechanism is given by:
\begin{equation}
	\mathrm{DLA}(\mathbf{f}, \mathbf{S}, \mathbf{A})=
	\sum_{m=1}^{M} \sum_{l=1}^{L}
	\sum_{p=1}^{P} \mathbf{A}_{mlp}\cdot \mathbf{F}_l( \phi(\mathbf{S}_{mlp})),
	\label{eq:mdla}
\end{equation}        
where $\mathbf{F} = \{\mathbf{f}_1, \ldots, \mathbf{f}_L\}$ is the set of feature maps, 
$\phi(\cdot)$ is a function that normalizes the coordinates of the sampling point $\mathbf{S}_{mlp}$ between $[-1, 1]$.
	
	\section{Results}
	\subsection{Datasets}
	The \textbf{ShangaiTech Wireframe dataset} \cite{wireframe_data}
	is a manually-labeled dataset for line detection consisting of  5462 images (indoor and outdoor). 
	The images contain man-made environments such as houses, bedrooms, kitchens, and living rooms. 
	The images provide lines with meaningful geometric information about the scene. We use it for training and validation, 
	splitting the data into 5000 and 462 images for each stage.
	
	The \textbf{York Urban dataset} \cite{york_data} is a manually-labeled dataset consisting of 122 images 
	(45 indoor and 57 outdoor). 
	We use this dataset for out-of-distribution testing.
	

	\subsection{Comparisons against other methods}
		\begin{table*}[!t]
		\centering
			\begin{tabular}{l c c c ccc p{0.2mm} ccc c}
				\toprule
				\multirow{2}{*}{Method} &  {\multirow{2}{*}{Epochs}} &  {\multirow{1}{*}{Params}} & {\multirow{1}{*}{FLOPS}} &\multicolumn{3}{c}{\textit{Wireframe Dataset} $\uparrow$} & \phantom{.} & \multicolumn{3}{c}{\textit{YorkUrban Dataset} $\uparrow$} & \multirow{1}{*}{Latency} \\
				\cmidrule{5-7} \cmidrule{9-11} 
				& & (M) $\downarrow$ & (G) $\downarrow$ & {sAP$^{5}$} & {sAP$^{10}$}  & {sAP$^{15}$}  & & {sAP$^{5}$}  & {sAP$^{10}$}  & {sAP$^{15}$} & (ms) $\downarrow$\\
				\midrule
				\multicolumn{10}{l}{CNN-based networks}\\
				L-CNN \cite{lcnn} & \underline{16} &  9.75 & 195.08 & 58.9 & 62.9 & 64.7 && 24.3 & 26.4  & 27.5 & 61.88\\
				HAWP \cite{hawp} & \underline{16} & 10.33 & 138.14 & 62.6 & 66.5 & 68.2 &&  26.2 & 28.6 & 29.7 & 32.92\\
				HAWPv2 \cite{HAWPv2} & 30 & 11.11 & 134.33  & 65.7 & 69.7 & 71.3 && 28.8 & 31.3 & 32.7 & 30.54\\
				F-Clip (HG1) \cite{fclip} & 300 & \underline{4.17} & 65.93 & 58.6 & 63.5 & 65.9 && 24.4 & 26.8 & 28.1 & 13.29\\
				F-Clip (HG2-LB) \cite{fclip} & 300 & 10.33 & 131.20 & 62.6 & 66.8 & 68.7 && 27.6 & 29.9 & 31.3 & 29.88\\
				F-Clip (HR) \cite{fclip} & 300 & 28.54 & 82.14 & 63.5 & 67.4 & 69.1 &&  28.5 & 30.8 & 32.2 & 63.75\\
				ULSD \cite{ulsd} & 30 & 9.99 & 114.17 & 65.0 & 68.8 & 70.4 & & 26.0 & 28.8 & 30.6 & {26.72}\\
				SACWP$^\dagger$ \cite{sacwp} & 30 & 10.2 & / & \underline{66.2} & \underline{70.0} & \underline{71.6} & & 27.2 & 30.0 & 31.8 & 28.74\\
				MLNET$^\dagger$ \cite{mlnet} & 30 & 16.85 & 75.05 & 65.1 & 69.1 & 70.8 & & 29.4 & 32.1 & 33.5 & 79.49\\
				\midrule
				\multicolumn{10}{l}{Transformer-based networks}\\
				LETR \cite{LETR} & 825 & 59.49 & 660.78 & 58.6 & 63.5 & 65.9 && 25.4 & 29.4 & 31.7 & 150.93\\
				DT-LSD \cite{dtlsd} & 24 & 217.16 & 1019.75 & \textbf{66.6} & \textbf{71.7} & \textbf{73.9} & & \underline{30.2} & \underline{33.2} & \underline{35.1}  & 176.23\\
				\textbf{LINEA-N} (ours) & 72 & \textbf{3.93} & \textbf{12.10} & 58.7 & 65.0 & 67.9 && 27.3 & 30.5 & 32.5 & \textbf{2.54}\\			
				\textbf{LINEA-S} (ours) & 36 & 8.57 & \underline{31.65} & 58.4 & 64.7 & 67.6 && 28.9 & 32.6 & 34.8 & \underline{3.08}\\			
				\textbf{LINEA-M} (ours) & 24 & 13.46 & 45.63 & 59.5 & 66.3 & 69.1 && 30.3 & 34.5 & 36.7 & 3.87\\			
				\textbf{LINEA-L} (ours) & \textbf{12} & 25.17 & 83.77 & 61.0 & 67.9 & 70.8 && \textbf{30.9} & \textbf{34.9} & \textbf{37.3} & 5.78\\	
				\bottomrule
			\end{tabular}
		\caption{Line detection results.
			Based on the models trained on the Wireframe dataset, we provide
			test results on both the YorkUrban and Wireframe datasets.
			The \textbf{best} results are given in boldface.
			Underlines are used for the $\underline{\text{second best}}$. We did not use the custom 
			cuda implementation for DT-LSD for deformable attention. $\dagger$ means the code is not available.
		}
		\label{tab: results}
	\end{table*}	
	
	\begin{figure}[!t]
		\centering
		\begin{tabular}{cc}
			\includegraphics[width=0.20\textwidth]{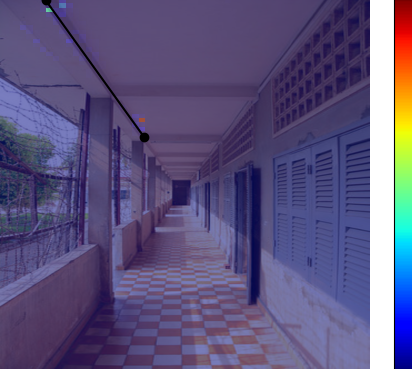} &
			\includegraphics[width=0.20\textwidth]{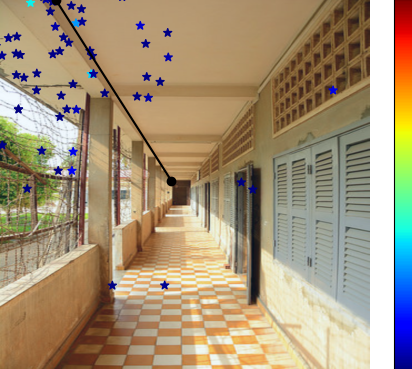} \\
			Attention \cite{attention} & Deformable attention \cite{deformable-DETR}\\
			\multicolumn{2}{c}{\includegraphics[width=0.20\textwidth]{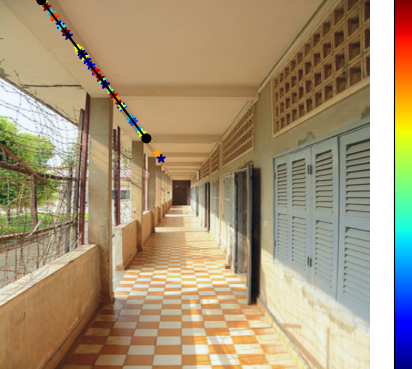}} \\
			\multicolumn{2}{c}{Deformable line attention (ours)}
		\end{tabular} 
		\caption{Attention mechanisms for line detection. We use a jet map to visualize where red means more important and blue is less important. For the sparse mechanisms, the stars represent the sampling points. The black line is the predicted line, and the black dots represent the line endpoints. 
		}
		\label{fig:attn_comparison}
	\end{figure}
	
		\begin{table}[!t]
		\resizebox{\columnwidth}{!}{
			\begin{tabular}{p{40mm} c ccc c}
				\toprule
				\multirow{2}{*}{Method} & \multirow{2}{*}{WsAP$^{10}$} & \multicolumn{3}{c}{\textit{YorkUrban Dataset}} & \multirow{1}{*}{FLOPS}\\
				& &  {sAP$^{5}$} & {sAP$^{10}$}  & {sAP$^{15}$}  & (G)\\
				\midrule
				D-FINE (baseline) \cite{dfine} & 37.7 & 10.4 & 17.7 & 21.7 & 95.02\\
				\phantom{a}$-$ DDL and FGL & 48.3  & 15.7 & 22.0 & 25.6 & 95.02\\
				\phantom{a}$+$ asymmetric convs & 45.9 & 15.9 & 22.5 & 25.7 & 95.02\\
				\phantom{a}$+$ replace MDA by DLA & 64.2 & 24.3 & 27.8 & 29.9 & 94.87\\
				\phantom{a}$+$ improved anchor generation (\cref{alg:anchor_generation}) & 64.6 & 24.6 & 28.5 & 30.5 & 94.87\\
				\phantom{a}$+$ new sampling points\phantom{text} (4, 1, 1) &  65.7 & 26.8 & 30.3 & 32.4 & 94.75\\
				\phantom{a}$+$ 1100 queries & 67.0 & 29.4 & 32.8 & 34.8 & 103.28\\
				\phantom{a}$+$ reduce hidden dimension in GELAN  & 67.2 & 29.6 & 33.3 & 35.7 & 83.77\\
				\phantom{a}$+$ new loss coefficients & 67.9 & 30.9 & 34.9 & 37.3 & 83.77\\
				\bottomrule
			\end{tabular}
		}
		\caption{Step-by-step progression from baseline model to LINEA-L. Each step shows the scores on YorkUrban, the FLOPS, and the sAP$^{10}$ on Wireframe (WsAP$^{10}$).}
		\label{tab:ablation}
	\end{table}
	
		\begin{figure*}[!t]
		\centering
		\begin{tabular}{cccc}
			\multicolumn{2}{c}{High-FLOPS models} & \multicolumn{2}{c}{Low-FLOPS models}\\
			\phantom{aaaaa}Wireframe & \phantom{aaaaa}YorkUrban & \phantom{aaaaa}Wireframe & \phantom{aaaaa}YorkUrban\\
			\includegraphics[width=0.23\textwidth]{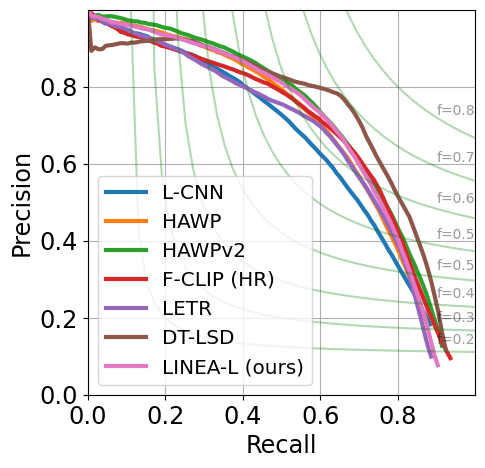} &  \includegraphics[width=0.23\textwidth]{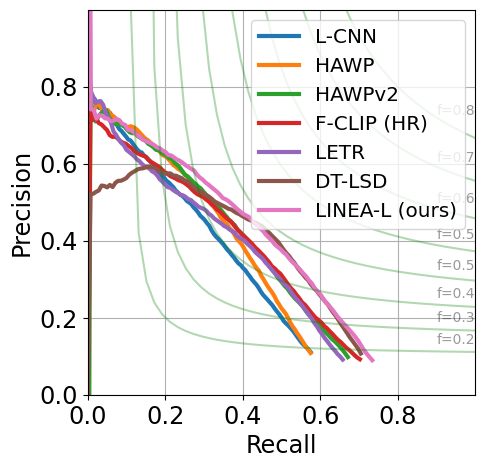} &
			\includegraphics[width=0.23\textwidth]{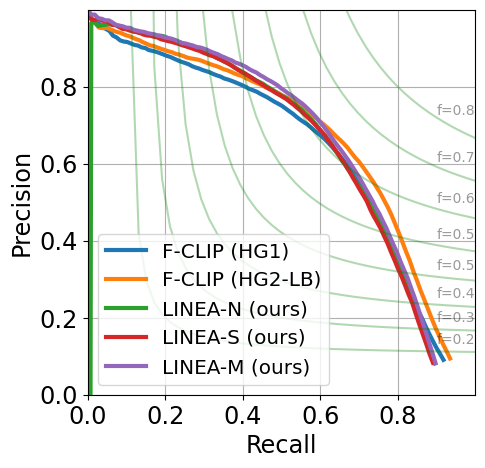} &  \includegraphics[width=0.23\textwidth]{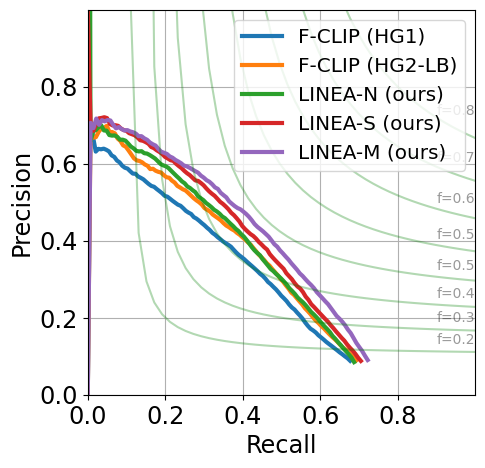} 
		\end{tabular}
		\caption{AUC curves for the sAP$^{10}$ on the Wireframe (left) and YorkUrban (right) datasets.
		}
		\label{fig:auc}
	\end{figure*}
	
	\begingroup
	\setlength{\tabcolsep}{1pt}
	\renewcommand{\arraystretch}{0.5}
	\begin{figure*}
		\resizebox{\textwidth}{!}{
			\begin{tabular}{c c c c c c c}
				\includegraphics[width=0.15\textwidth]{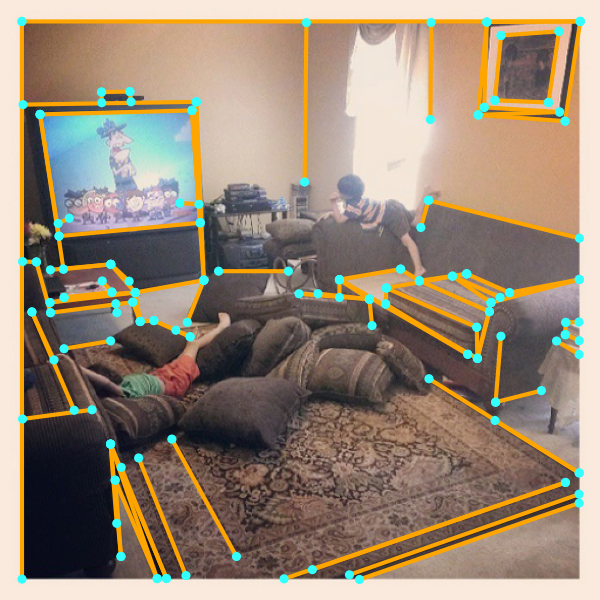}&\includegraphics[width=0.15\textwidth]{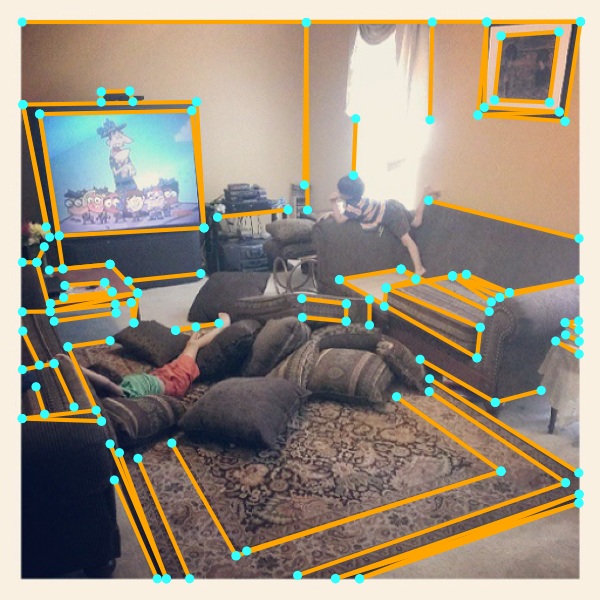}&\includegraphics[width=0.15\textwidth]{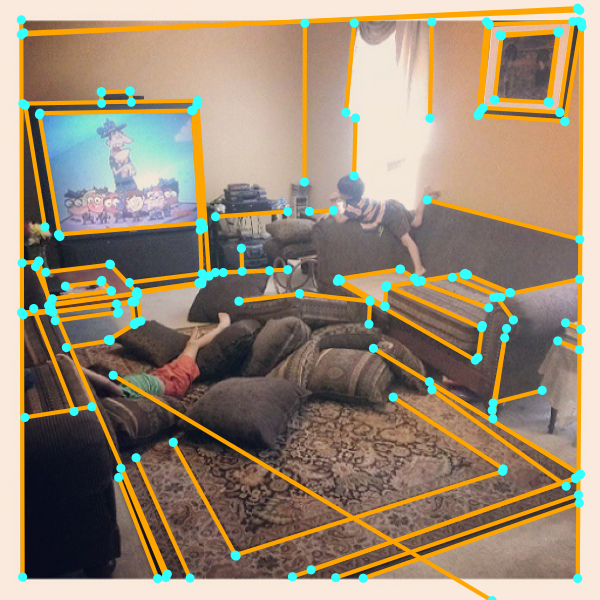}&\includegraphics[width=0.15\textwidth]{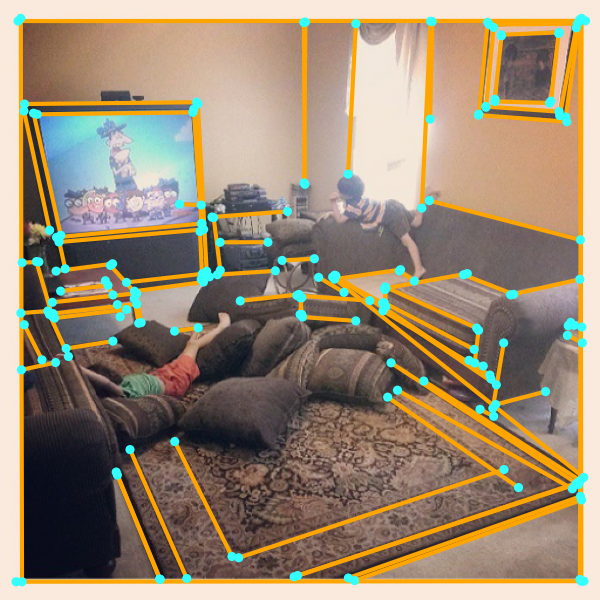}&\includegraphics[width=0.15\textwidth]{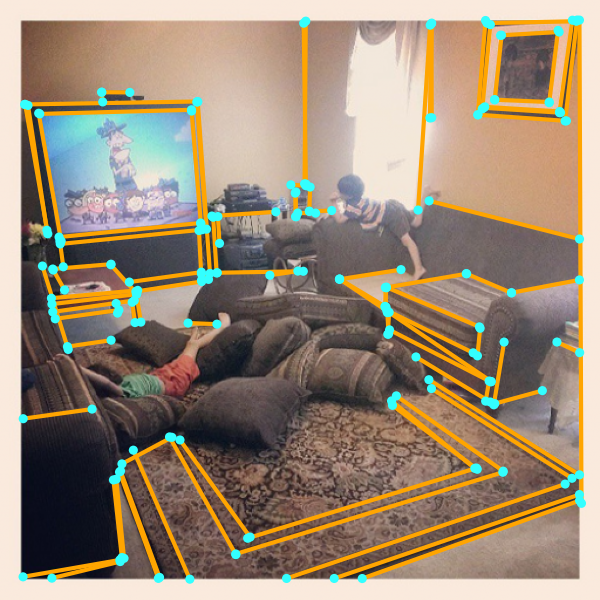}&\includegraphics[width=0.15\textwidth]{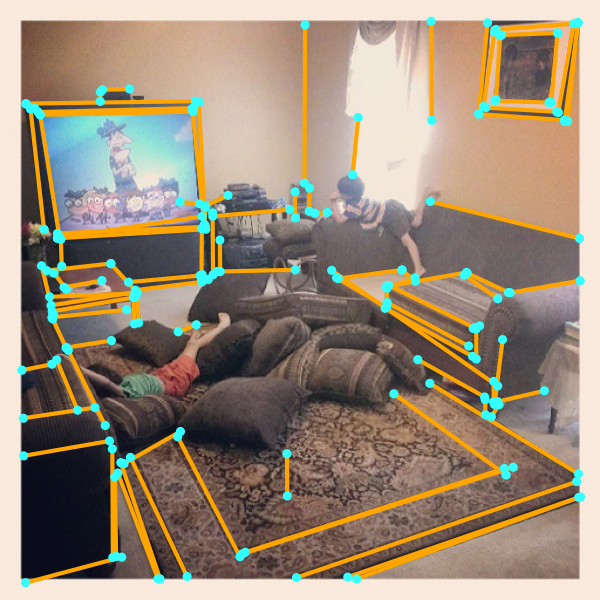}&\includegraphics[width=0.15\textwidth]{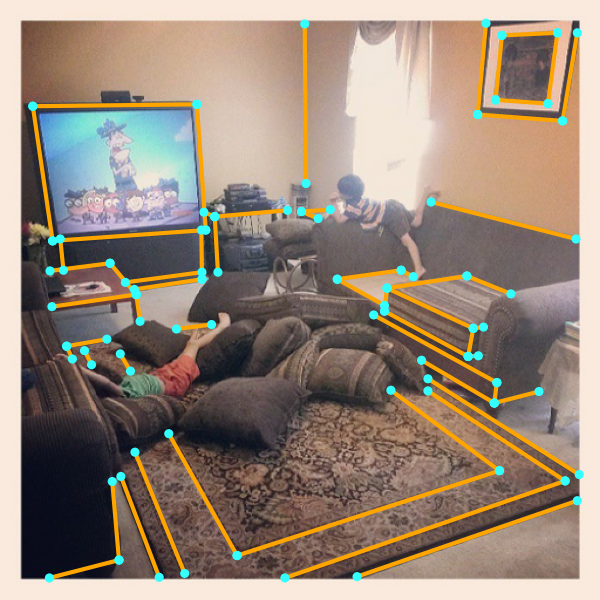}\\
				\includegraphics[width=0.15\textwidth]{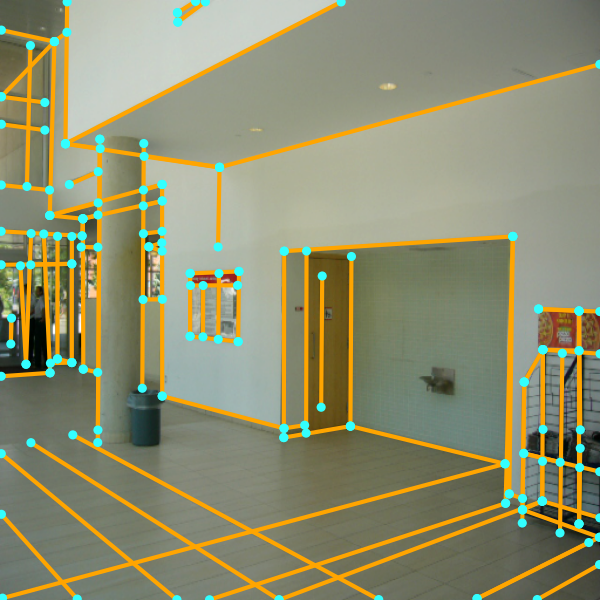}&\includegraphics[width=0.15\textwidth]{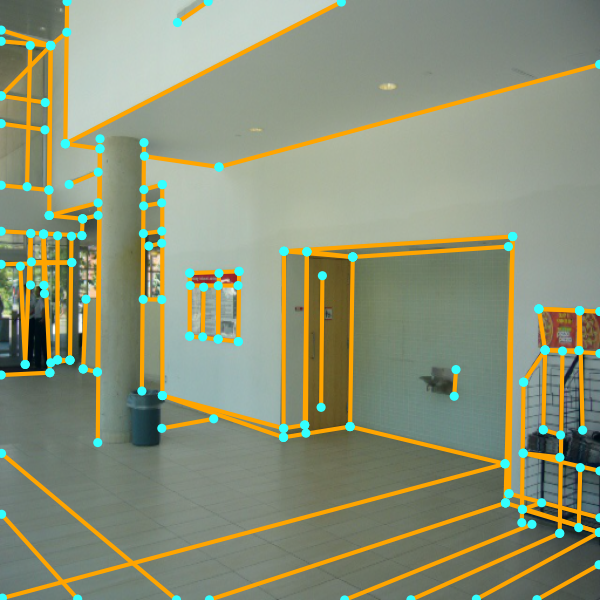}&\includegraphics[width=0.15\textwidth]{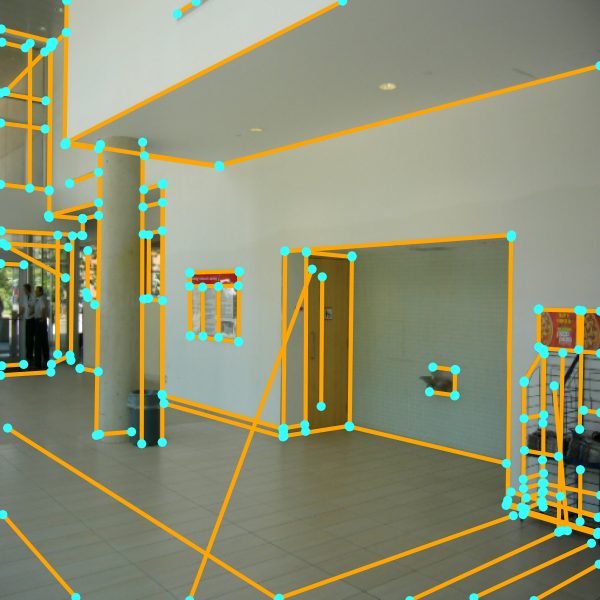}&\includegraphics[width=0.15\textwidth]{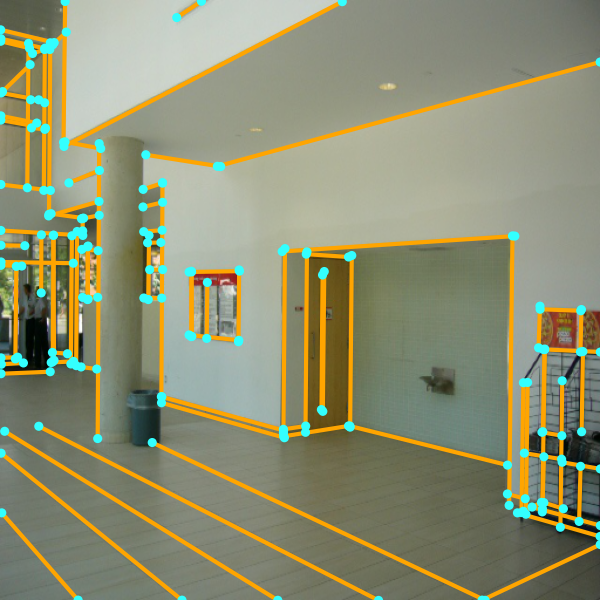}&\includegraphics[width=0.15\textwidth]{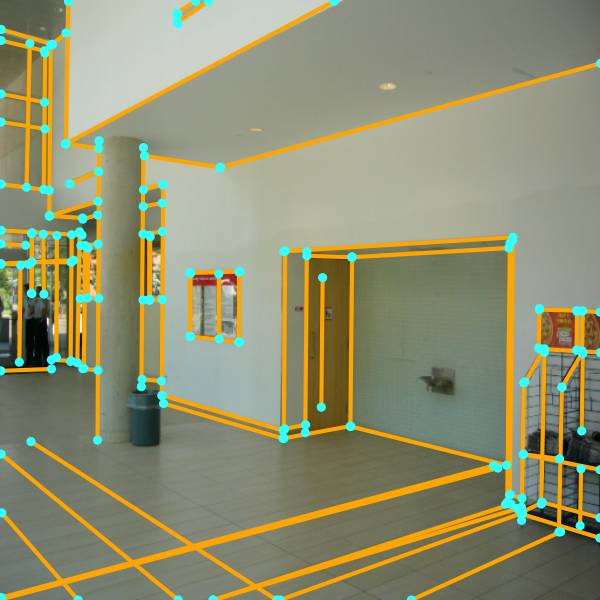}&\includegraphics[width=0.15\textwidth]{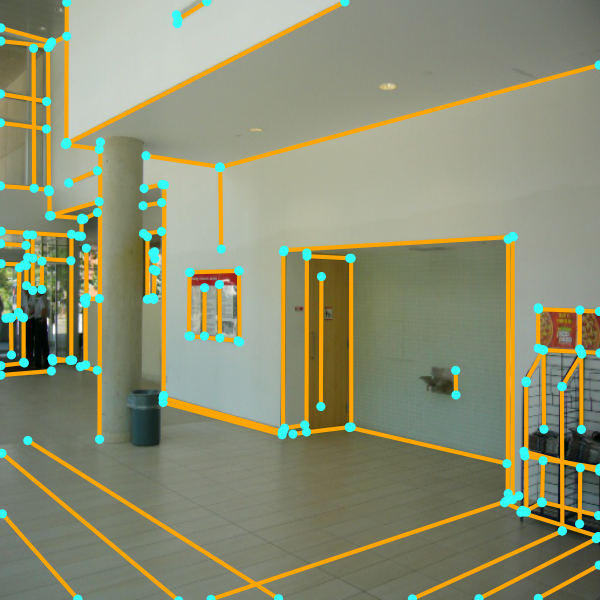}&\includegraphics[width=0.15\textwidth]{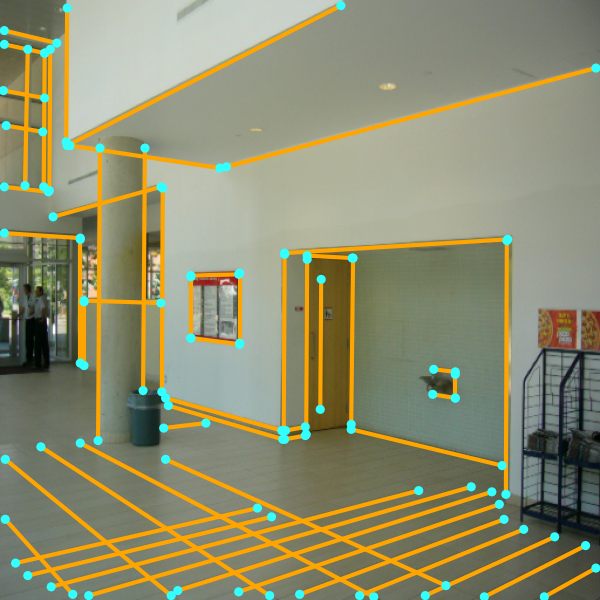} \vspace{3pt}\\
				L-CNN \cite{lcnn} & HAWPv2\cite{hawp} & F-CLIP (HR) \cite{fclip} & LETR \cite{LETR} & DT-LSD \cite{dtlsd} & LINEA-L (ours) & Ground Truth
			\end{tabular}
		}
		\caption{Line segment detection examples from Wireframe dataset (top) and the YorkUrban dataset (bottom).}
		\label{fig:results}
	\end{figure*}
	\endgroup
	By varying the parameters (e.g., the backbone and GELAN), we generate four models: LINEA-N, LINEA-S, LINEA-M, and LINEA-L. 
	We apply TensorRT FP16 to implement our LINEA models. 
	\Cref{tab: results} shows comparative results for several methods.
	We run all open-source models on an NVIDIA RTX A5500 with 24GB. Refer to \cref{tab:hyperparameter} for information about the training and LINEA models hyperparameters.
	
	All models use Wireframe for training. 
	We apply transformer-based data augmentation practice by modifying images without increasing the training dataset. 
	For CNN models, datasets sizes are increased by $4\times$.  
	For the initialization weights, LINEA models use pre-trained weights from CIFAR10 for the backbone.
	This is in contrast to other transformer models that require pre-training of all their components on a larger dataset 
	        ($>90$k images) for at least 20 epochs.
	
	
	In terms of inference latency, our proposed models are the fastest by significant margins.
	In terms of complexity, our LINEA models require less FLOPS than all other models. 
	Furthermore, in terms of out of distribution testing, our LINEA-L gave the best performance for all metrics.
	During training, LINEA-L also converge faster than any other model at just 12 epochs.
	Among parameter models, LINEA-L, uses a significantly smaller number of parameters than any other tranformer model (excluding other LINEA variations).
	
	Our accuracy results for the Wireframe validation set are competitive.
	Among all models, LINEA-L provides the third highest sAP$^5$ metric.
	Nevertheless, in terms of generalization performance,
	we note that it is more significant that we outperform all other models in all metrics in the out-of-distribution testing
	on the YorkUrban dataset.
	
	For a further analysis, we show the area under the curve (AUC) for the sAP$^{10}$ in \cref{fig:auc}. 
	In the high-FLOPS models category, LINEA-L produces low precision values for high recall, 
	    but performs well for low recalls ($<0.4$); second to 
	HAWPv2 \cite{HAWPv2} for the wireframe dataset. 
	The AUC for the YorkUrban dataset shows that LINEA-L has a better performance than all the other models. 
	For low-FLOPS models, LINEA-N surpasses all F-CLIP models on the YorkUrban datasets while being the smallest model in \cref{tab: results}. On the Wireframe data, LINEA-N, -S and -M have high precision for low recall values. We also provide example results images for the two datasets in \cref{fig:results}. For a fair visual comparison, we plot the first 100 lines with the highest probabilities for each model.
	
	For a single-line detection example, we compare several attention mechanisms in \cref{fig:attn_comparison}. 
	The example clearly demonstrates that our proposed deformable line attention mechanism correctly samples
	several point along the line.
	In contrast, the original attention mechanism \cite{attention} attempt to sample the entire map while
	DT-LSD \cite{dtlsd} misses several samples along the desired line.
	
	\subsection{Ablation study}
	\Cref{tab:ablation} shows the step-wise progression from the baseline D-FINE model to LINEA-L. 
	The results from \cref{tab:ablation} demonstrate that our proposed components are essential for achieving high performance.
	
	 The addition of line contrastive denoising from DT-LSD to the baseline model does not work well for line detection.
	 Removing the DDL and FGL losses leads to significant gains, as these losses depend on the intersection-over-union metric, 
	   which cannot be computed with line instances. 
	 By adding $1\times3$ and $3\times1$ convolutions to the GELAN module, we observe slight score improvements on YorkUrban, despite a decrease in WsAP$^{10}$. This addition has no impact on FLOPS due to the fusion of convolutional kernels during deployment. Next, we replace MDA with DLA as the cross-attention mechanism, resulting in improved metrics and a $0.15$G reduction in FLOPS. The sAP$^{5}$ score increases by 8.4 percentage points. We also shift from bounding box to line representation for anchor initialization and adjust the sampling points from $(4, 4, 4)$ to $(4, 1, 1)$ to support line detection. Increasing the number of queries from 500 to 1100 boosts scores, though it significantly raises FLOPS. We mitigate this by reducing GELAN dimensions cutting FLOPS by about 20\% while enhancing metrics. Finally, we increase the classification loss coefficient from 1 to 4.
		
	\section{Conclusion}
	In this paper, we present a new family of line detectors based on end-to-end transformers.
	Our proposed models outperform all other line detection methods in terms of speed and out-of-distribution
	test accuracy.
	
	\section{Acknowledgment}
	This work was supported in part by the National Science Foundation under Grant 1949230.
	
	
	\bibliographystyle{IEEEbib}
	\bibliography{main.bib}

	\newpage
	
	\onecolumn
	\appendix
	
	\section{Hyperparameter configuration}
	\Cref{tab:hyperparameter} summarizes the hyperparameters for the LINEA models. All variants use HGNetv2 backbones pretrained on ImageNet \cite{imagenet22k} and the AdamW optimizer. During training, we apply the following data augmentation techniques: $\mathrm{HorizontalFlipping}$, $\mathrm{VerticalFlipping}$, $\mathrm{ColorJittering}$, $\mathrm{RandomCropping}$ and $\mathrm{RandomResizing}$.
	\begin{table*}[h]
		\centering
		\begin{tabular}{l | c c c c}
			\toprule
			Setting & LINEA-L & LINEA-M & LINEA-S & LINEA-N\\
			\midrule
			Backbone Name & HGNetv2-B4 & HGNetv2-B2 & HGNetv2-B0 & HGNetv2-B0\\
			Optimizer & AdamW & AdamW & AdamW & AdamW\\
			Embedding Dimension & 256 & 256 & 256 & 128 \\
			Feedforward Dimension & 1024 & 512 & 512 & 512 \\
			GELAN Hidden Dimension & 64 & 42 & 42 & 22\\
			GELAN Depth & 3 & 3 & 2 & 2\\
			Decoder Layers & 6 & 4 & 3 & 3\\
			Number of Queries & 1100 & 1100 & 1100 & 1100\\
			Bin Number & 16 & 16 & 16 & 16\\
			Sampling Point Number & (4, 1, 1) &  (4, 1, 1) &  (4, 1, 1) &  (4, 1, 1)\\
			Base Learning Rate & 2.5e-4 & 2e-4 & 2e-4 & 8e-4 \\
			Backbone Learning Rate & 1.25e-5 & 2e-5 & 1e-4 & 4e-4\\
			Weight Decay & 1.25e-4 & 1e-4 & 1e-4 & 1e-4\\
			Weight of $\mathcal{L}_{\text{line}}$ & 5 & 5 & 5 & 5\\
			Weight of $\mathcal{L}_{\text{class}}$ & 4 & 1 & 1 & 1\\
			Total Batch Size & 8 & 8 & 8 & 8\\
			Epochs & 12 & 24 & 36 & 72\\
			\bottomrule
		\end{tabular}
		\caption{Hyperparameter configuration for different LINEA modes.}
		\label{tab:hyperparameter}
	\end{table*}
	
\end{document}